# Exploring the Effectiveness of Methods for Persona Extraction

Konstantin Zaitsev [0009-0001-1389-5865]

Higher School of Economics, Moscow, Russia
kzaytsev@hse.ru

**Abstract.** The paper presents a study of methods for extracting information about dialogue participants and evaluating their performance in Russian. To train models for this task, the Multi-Session Chat dataset was translated into Russian using multiple translation models, resulting in improved data quality. A metric based on the F-score concept is presented to evaluate the effectiveness of the extraction models. The metric uses a trained classifier to identify the dialogue participant to whom the persona belongs. Experiments were conducted on MBart, FRED-T5, Starling-7B, which is based on the Mistral, and Encoder2Encoder models. The results demonstrated that all models exhibited an insufficient level of recall in the persona extraction task. The incorporation of the NCE Loss improved the model's precision at the expense of its recall. Furthermore, increasing the model's size led to enhanced extraction of personas.

**Keywords:** Persona Extraction, Dialogue Datasets, Seq2Seq Models, Translation Dataset, Persona Matching

## 1 Introduction

Modern language models can conduct conversations as chatbots with users. Conversations can involve exchanging information about the user's life, such as education, work, preferences, relationships and so on. This information is known as personas. Personas can influence on the engagement in the conversation and make communication with the chatbot feel more natural [1, 2, 3].

An important aspect of interaction with a chatbot is the ability to identify, or extract, personas in the communication process. However, there is a lack of attention in the academic literature on the methods for persona extraction and evaluating their effectiveness. Most research focuses on using personas in conversations, but their extraction methods are often not mentioned or not described clearly. Therefore, it is crucial to investigate which approaches can provide the most efficient extraction and what challenges models face.

Most of the research primarily concentrates on dialogue models and data in English, which leads to the ignorance of other languages. To tackle this issue for the Russian language, this study aims to adapt the dialogue dataset with personas and models based on English-language resources. The techniques presented in the study are expected to be applicable to other languages as well.

The objective of this paper is to identify efficient methods for extracting personas from dialogues in Russian. To achieve this goal, we focus on two key aspects: adapting dialogue data into Russian and evaluating the effectiveness of persona extraction techniques. The research contribution is as follows:

1. Translation of the English dialog dataset into Russian. The paper describes the process of creating Russian-language data from the source language dataset. The presented approach can be applied for other languages.
2. Persona Extraction Metric. The research proposes a comprehensive evaluation framework to assess the quality of persona extraction models. This includes developing metrics based on embeddings and a matching process of target and extracted personas, which measures the precision and recall of persona extraction models.
3. Identifying Weaknesses of Persona Extraction Models. The results of experiments on various parameters and training methods demonstrate which techniques are effective in persona extraction, as well as the challenges faced by the models.

## 2 Related Works

The summarization and persona extraction tasks are typically addressed using models based on the Sequence-to-Sequence (Seq2Seq) transformer architecture [4]. Some of popular models include BART [5] and T5 [6]. For these models there are multilingual variants that support the Russian language. These include MBart [7], mT5 [8] and mT0 [9]. The latter is a finetuned mT5 on the cross-lingual task mixture. Also Russian is supported by models, such as FRED-T5 and ruT5 [10], specifically pretrained on Russian and English corpora.

In addition to Seq2Seq models, large language models (LLM) based on a decoder architecture can also be used. Especially, their use within the framework of retrieval-augmented generation (RAG) has become particularly popular [11, 12]. However, the high computational costs associated with their application present a significant challenge, potentially limiting their applicability within the production pipeline.

The persona extraction task is similar to the dialogue summarization. In summary, it is important to convey the key points of the conversation. The extracted bulletpoints may also contain personas. There are models for summarizing dialogues, but most of them have only been trained on English language datasets.

Several English dialogue datasets are available, including Persona Chat [13] and Multi-Session Chat (MSC) [14], which both contain information about the participants. The MSC dataset is split into multiple sessions to simulate interrupted communication. Additionally, SamSum [15] and DialogSum [16] are used for summarizing dialogues.

The described datasets include dialogues in English only. Translated versions of SamSum[1] and DialogSum[2] in Russian are also available. There is a Matreshka dataset[3] based on synthetic dialogues generated by ChatGPT. The dataset contains personas and summaries. However, a significant issue with the dataset is the low quality of the generated personas and the unnatural communication. In addition to the mentioned datasets, Toloka crowdsourced a more natural dialogue dataset that includes personas[4].

It is common for data related to a task to be limited to a specific language, hindering model development for other languages. To address this limitation, we explore methods for adapting models to new languages, focusing on the approach of training a model on a translated dataset.

This approach involves translating the training dataset into the target language before training the model. This method has the advantage of reducing the total number of required inferences to one, eliminating the latency issue associated with other approaches. Notably, our approach has been inspired by the success of similar methods in text detoxification tasks using multilingual models [17], where training on translated data achieved results comparable to those of models fine-tuned on monolingual datasets.

## 3  Methods

In order to extract personas from Russian language dialogues, it is necessary to develop models fine-tuned on data for this task. To accomplish this, the MSC must first be translated into Russian using a heuristic process that will be described in detail below. To assess the quality of persona extraction, a metric based on F-measure is proposed. This metric allows for the evaluation of the precision and recall of persona extraction. The metric results can help identify shortcomings in persona extraction models.

### 3.1  Dataset Translation

The translation process could be utilized by commercial and opensource translators. Currently, commercial translators produce high-quality translations. However, their usage in translating entire datasets might be expensive. Inference of opensource models is much cheaper than usage of commercial translators. Moreover, in the majority of cases, opensource models could translate the utterance correctly. Conversely, opensource models could sometimes generate phrases that lose important information or make the translation meaningless. Consequently, the quality of persona extraction could not be optimal. Therefore, there is a need to combine commercial and opensource translators.

Among the described persona datasets, experiments were conducted on the MSC. This choice is due to the ease of data collection for training models, as well as the availability of long dialog contexts in the MSC corpus. In order to finetune the Russian model, it was necessary to translate the MSC data from English. There are many models that can be used for the translation task. This study uses a combination of the NLLB model [18] and the Yandex Translator[5]. The models have demonstrated satisfactory performance in benchmarking and experimental studies, rendering them a suitable choice for our intended purposes.

---

[1] https://huggingface.co/datasets/d0rj/samsum-ru
[2] https://huggingface.co/datasets/rcp-meetings/rudialogsum_v2
[3] https://huggingface.co/datasets/zjkarina/matreshka
[4] https://www.kaggle.com/datasets/valentinbiryukov/toloka-persona-chat-rus

NLLB can only translate sentences, which can lead to limitations when translating entire dialogues. Specifically, when translating sentences within utterances or the utterances themselves, some contextual information may be lost, resulting in incomplete translations and potential issues. For instance, translations of sentences may not accurately preserve the relationships between utterances, leading to inconsistencies in gender or pronoun usage.

However, despite these limitations, we found that the translated dialogues were still suitable for persona extraction. This is because the extracted personas are based on the overall characteristics and traits expressed in the dialogue, rather than relying solely on pronoun usage. While pronouns can provide important cues for identifying speakers, they are not the only indicators of persona. The NLLB translations, although imperfect, still capture the essential information needed for persona extraction.

The translation process has also revealed instances of agrammatical or incorrect translations that required correction. These cases could be identified by calculating perplexity or by classifying sentences according to their grammatical structure. Calculating perplexity requires a model that provides probabilities for output tokens. As the language model's inference is used in this calculation, this approach requires significant computational and time resources.

Instead, a grammar classifier trained on a manually annotated RuCOLA dataset [19] was used. Among the trained classification models, RoBERTa [20] was chosen. The model demonstrated a high performance on the benchmark[6].

To filter the data, we have developed an algorithm that assesses the grammatical accuracy of an utterance. Each utterance or personas sample is split into sentences, which are then subjected to grammatical analysis. For sentences, the grammar classifier is applied, which predicts the probability of grammaticality of the sentence. If the probability of one of the sentences is below a certain threshold the utterance or personas sample is considered corrupted. Subsequently, these corrupted utterances or personas are then translated using Yandex Translate to reduce the number of poorly translated phrases in the dataset. Table 1 presents statistics on well-translated utterances or personas. The dataset is accessible via the HuggingFace platform[7].

**Table 1.** Corrupted translations statistics

|  | All | Corrupted | Ratio |
|---|---|---|---|
| Utterances | 183124 | 61194 | 0.33 |
| Personas | 116404 | 11620 | 0.10 |

---

[5] https://translate.yandex.ru
[6] https://rucola-benchmark.com/leaderboard
[7] https://huggingface.co/datasets/adugeen/RuTranslatedMultiSessionChat

## 3.2 Persona Extraction Metric

The existing metrics lack the capacity to provide an indication of the extent to which the extracted personas are complete and accurate. Additionally, the calculation of these metrics is challenging due to the potential for the model to generate personas in a different order than that specified in the target. This can result in an overestimation of the quality of the model. Therefore, there is a need to introduce a new metric for persona extraction.

**Persona Classifier.**

To evaluate the quality of persona extraction, we suggest using a classifier and sentence encoder model to compare the predicted personas with the target ones. The use of the classifier is necessary for several reasons. Firstly, there may be instances when a particular persona is missing from the target data, for example, if a person was described in a prior session and is therefore not present in the current session. However, the model correctly extracted the persona. Secondly, the classifier enables the filtering out of redundant personas that were mistakenly identified by the extraction model. By applying a suitable threshold and ranking system, it becomes possible to select the most relevant participants in the dialogue.

The task of classification is to determine whether the extracted persona belongs to one of the two participants in the dialogue. The dataset contains only two people, so we used two classes in the training process: to simulate cases where the extracted persona might not be relevant to either of the dialogue participants, we added a third class called 'none'. Therefore, the classifier must predict one of three possible targets: 'bot_0', 'bot_1', or 'neutral'.

The dataset was created by selecting a persona related to a specific participant from a list of available personas for each dialogue. For the neutral category, personas from other dialogues were randomly chosen. The training data consists of dialogues, personas, and the target class indicating the dialogue participant. The label distribution in the dataset was uniform across all categories. An example from the dataset is included in Appendix A.

To train a classifier, the most suitable model is a pretrained model on the Natural Language Inference (NLI) task. The classification of personas can be formulated as a task of determining the relationship between dialogue, participant, and persona (entailment, non-entailment, or neutral). For finetuning, a multilingual model based on RoBERTa architecture[8] was selected. The classifier was finetuned on an English language dataset in order to assess the transferability of knowledge from English to Russian. The results of the classification based on test data in Russian and English are presented in Table 2. The average F1 score for Russian and English datasets is 0.81 and 0.9 respectively.

---

[8] https://huggingface.co/MoritzLaurer/multilingual-MiniLMv2-L6-mnli-xnli

Table 2. Classification metrics. The label "Neutral" means that the persona is not relevant for both participants

| Language | Label | Precision | Recall | F1-score | Support |
|---|---|---|---|---|---|
| Russian | Bot_0 | 0.79 | 0.80 | 0.79 | 12769 |
|  | Bot_1 | 0.76 | 0.83 | 0.79 | 13550 |
|  | Neutral | 0.90 | 0.80 | 0.84 | 12739 |
| English | Bot_0 | 0.89 | 0.87 | 0.88 | 12951 |
|  | Bot_1 | 0.88 | 0.88 | 0.88 | 13519 |
|  | Neutral | 0.92 | 0.94 | 0.93 | 12838 |

**Persona Matching.**

The predicted personas may differ in spelling from those of the targets, but the meaning will remain similar. Furthermore, the order in which the predicted personas are presented may differ from that presented in the target set. As a result, the metric computation could be inaccurate and may underestimate the model's accuracy.

Various techniques can be used to convert text data into numerical representations. These approaches can range from simple methods such as tf-idf to more resource-demanding models such as sentence encoders. To ensure a high-quality comparison between the extracted and target personas, the E5 model was employed [21] as it has demonstrated a high level of performance based on the results of the MTEB[9].

The matching algorithm consists of several steps. Firstly, the targets and predicted personas are segmented into sentences and placed into separate lists. Next, the embeddings of the sentences in each list are calculated. The similarity between the vector representations of the extracted and targeted personas is then determined using cosine distance. To account for identical personas, a threshold value is applied.

**Metrics Calculation.**

Precision, recall, and F-score were calculated using the resulting matches. Precision was determined by the ratio of correctly identified extracted personas to the total number of extracted personas. Recall, on the other hand, was determined by the ratio of correctly matched personas from the list of true personas to the total number of true personas.

Additionally, the metrics also accounted for cases where the extracted persona was not directly related to the target but was instead associated with a participant in the dialogue. These cases slightly underestimated the metrics, resulting in an inaccurate reflection of the models' quality. To address this issue, we have devised a method that utilises a persona classification model.

If the extracted persona has a similarity score lower than the threshold, we calculate its probability of being assigned to a specific participant in the conversation using the classification model described earlier. Once a certain level of confidence is reached, the persona is considered to successfully extracted. This process improves the precision and recall of the persona extraction task. To measure the effectiveness of our approach, we use the following metrics.

---

[9] https://huggingface.co/spaces/mteb/leaderboard

The precision metric calculates the proportion of extracted personas that are correct and relevant. It is defined as:

$$Precision = \frac{Matched\ Extracted\ Personas + Classified\ Personas}{Total\ Extracted\ Personas} \qquad (1)$$

Here, *Matched Extracted Personas* refers to the number of personas that were correctly identified, *Classified Personas* denotes the number of personas whose probability of matching a persona to the required participant in the dialogue is above the threshold, and *Total Extracted Personas* includes all extracted personas.

The recall metric measures the proportion of relevant personas that were successfully matched. It is defined as:

$$Recall = \frac{Matched\ Target\ Personas + Classified\ Personas}{Total\ Target\ Personas + Classified\ Personas} \qquad (2)$$

In this formula, *Matched Target Personas* represents the number of personas from the list of all target personas that were matched with the extracted personas, *Classified Personas* is the same as in the precision metric, and *Total Target Personas* corresponds to the list of all target personas.

The F1 score is a harmonic mean of precision and recall, providing a balanced measure of both metrics. It is calculated as:

$$F1 = 2 \cdot \frac{Precision \cdot Recall}{Precision + Recall} \qquad (3)$$

In order to further assess the quality of persona extraction and to compare a proposed metric with the common metrics used in text summarization tasks, the Rouge [22], BLEU [23], METEOR [24] and BERTScore [25] metrics were also utilized.

## 4    Experiments

The study analyses the effectiveness of different approaches to training models for extracting personas. The objective of all experiments was to generate personas based on the dialogue. For each dialogue participant, a prompt was provided that specified for whom personas needed to be extracted. A prompt format is presented in Appendix B.

For some of the models, finetuning was conducted on both the translated dataset and the combination of translated and original English dialogues. The rationale behind this approach was that the presence of samples in both languages could enhance cross-language knowledge transfer. Table 3 displays the size of the train and test sets.

Experiments were conducted for MBart-large, FRED-T5-large, Starling-7B-beta [26], which is based on Mistral-7B-Instruct [27], ruT5 and mT0-large. The selection of the model was driven by two key considerations: firstly, the exploration of the multilingual abilities of cross-language knowledge transfer, and secondly, the investigation of the impact of model size on the quality of persona extraction.

Table 3. Sample sizes for finetuning of persona extraction models

| Language | Train | Test |
|---|---|---|
| Russian | 22862 | 5716 |
| English | 22862 | 5716 |

### 4.1 Finetuning Pretrained Model on Similar Tasks

It is hypothesized that a model pretrained on text summarization tasks may yield better results than simple model finetuning. Additionally, pretraining on the task of machine translation, as proposed in the task of text detoxification, may lead to quality improvement. The assumption is that translation will assist the model in transferring knowledge from English to Russian more effectively.

Datasets such as DialogSum[10] and SamSum[11] were used for the summarization task. Besides, the Russian-translated versions[12,13] were also used. For the machine translation task, the news commentary[14] and opus-100 [28] data were used.

The pretraining process consisted of five epochs, and the final model was selected based on the checkpoint with the lowest loss function value. Subsequently, the resulting model was further trained on the translated dataset, following the same procedure as in the previous section.

### 4.2 Training with NCE loss

To enhance persona extraction quality, we incorporated the Noise-Contrastive Estimation (NCE) Loss [29] alongside the Cross-Entropy Loss. The NCE loss serves to increase the distance between positive and negative embeddings. In the context of persona extraction, positive examples refer to target personas, while negative examples pertain to personas belonging to the other dialogue participant.

It is hypothesized that minimizing the NCE Loss may improve the precision of a persona extractor. This means that the model is expected to generate relevant personas more frequently. The NCE was combined with cross-entropy. Equation 4 illustrates the manner in which losses are combined.

$$Loss = \alpha \cdot NCE + \beta \cdot CE \qquad (4)$$

In the equation *NCE* represents Noise-Contrastive Estimation, while *CE* stands for Cross-Entropy. Experimental evidence indicates that the equal values for $\alpha$ and $\beta$ are optimal for the finetuning. Consequently, the following weights were employed: $\alpha = 1$ and $\beta = 1$.

---

[10] https://huggingface.co/datasets/knkarthick/dialogsum
[11] https://huggingface.co/datasets/samsum
[12] https://huggingface.co/datasets/d0rj/dialogsum-ru
[13] https://huggingface.co/datasets/d0rj/samsum-ru
[14] https://huggingface.co/datasets/Helsinki-NLP/news_commentary

### 4.3 Encoder2Encoder Finetuning

A hypothesis was formed regarding the possibility of developing a model that combines encoder models as both an encoder and a decoder, based on the findings from experiments with the training of a persona classifier. Previous studies have shown promise for this approach [30]. In this study, the model was developed based on a relevance classification model for personas. The encoders and decoders' embeddings were shared, as this approach is expected to produce high-quality results based on the mentioned research.

Another approach, similar to abstractive summarization, can be employed in addition to constructing a Seq2Seq model using encoders. This approach, called extractive summarization, involves highlighting a piece of text related to the persona of a specific participant from the dialogue. However, this method presents challenges as it requires matching and aligning specific personas with the text. Therefore, in this work, we only used Seq2Seq for persona generation.

### 4.4 Large Models Finetuning

The next stage of the research aims to evaluate the quality of training larger models. For this purpose, the FRED-T5 model was employed, which was trained on a translated dataset without pretraining on similar persona extraction tasks. In addition to the selected model, Starling-7B was considered due to its high performance on the LMSYS leaderboard[15]. Experiments were conducted to compare the results of fine-tuning the lighter models, including MBart, mT0 and ruT5.

The FRED-T5 model was trained using NCE, which involved both positive and negative personas. This was done to prove the hypothesis that using NCE loss can improve precision. Due to the significant computing resources required for training of FRED-T5 with NCE Loss and Starling-7B, LoRA [31] was used to train a portion of the weights while maintaining a similar level of quality as a full fine-tuning process. The training utilized the following parameters: r=16, lora_alpha=32, and lora_dropout=0.1.

### 4.5 Finetuning on Translated and Original Dialogues

The transfer of cross-lingual knowledge could be supported by finetuning not only on similar tasks but also on translated and original samples. To explore the impact of such an approach, a dataset was collected that contains translated into Russian and original English dialogues. The underlying hypothesis is that if the model is able to solve the task in both languages, the quality of persona extraction would be higher as the knowledge is transferred between languages.

To prove this hypothesis, mT0, Starling-7B and ruT5 were selected. As mT0 was finetuned on a diverse range of cross-lingual tasks, it is anticipated that the model may yield superior results compared to MBart. Starling-7B was selected to demonstrate that the quality achieved through finetuning on both languages and size could potentially exceed that of other configurations. Finally, experiments with ruT5 are necessary to assess the impact of finetuning a monolingual model. The aforementioned model was therefore subjected to finetuning on solely translated Russian dialogues.

---

[15] https://arena.lmsys.org

## 5   Results

**Table 4.** The results of various approaches to training models for persona extraction. Here, P, R, F1 represent, respectively, Precision, Recall and F1 for persona extraction. Metrics with the "bert" prefix correspond to the bertscore Precision, Recall and F1 measures. MBart is a simple finetuned MBart model for persona extraction. Pre-MBart is pre-trained MBart fine-tuned on the persona extraction task. NCE-Pre-MBart is Pre-MBart with the additional NCE Loss. Enc2Enc is the Seq2Seq model based on encoders, trained on the persona classification task. The NCE-FRED-T5 is identical to the FRED-T5, except for the addition of an NCE Loss. Models bearing the "ru" and "ru-en" designations are subjected to fine-tuning on translated dialogues alone, as well as on translated and original dialogues.

| Model | P | R | F1↓ | $P_{bert}$ | $R_{bert}$ | $F1_{bert}$ | rougeL | bleu | meteor |
|---|---|---|---|---|---|---|---|---|---|
| Starling-7B-ru-en | **0.902** | 0.749 | **0.818** | 0.837 | 0.806 | **0.820** | **0.495** | **0.289** | **0.462** |
| Starling-7B-ru | 0.897 | 0.744 | 0.813 | 0.834 | 0.803 | 0.817 | 0.488 | 0.284 | 0.457 |
| FRED-T5 | 0.879 | 0.753 | 0.811 | 0.829 | 0.810 | 0.818 | 0.483 | 0.283 | 0.462 |
| Pre-MBart | 0.864 | 0.755 | 0.805 | 0.816 | 0.810 | 0.812 | 0.462 | 0.272 | 0.452 |
| MBart | 0.860 | **0.756** | 0.805 | 0.820 | 0.811 | 0.814 | 0.467 | 0.271 | 0.454 |
| NCE-Pre-MBart | 0.869 | 0.742 | 0.801 | 0.821 | 0.808 | 0.813 | 0.468 | 0.267 | 0.448 |
| mT0-large-ru-en | 0.866 | 0.724 | 0.789 | 0.821 | 0.808 | 0.813 | 0.463 | 0.254 | 0.437 |
| ruT5 | 0.867 | 0.716 | 0.784 | 0.827 | 0.802 | 0.813 | 0.470 | 0.251 | 0.433 |
| NCE-FRED-T5 | 0.889 | 0.677 | 0.769 | 0.830 | 0.790 | 0.808 | 0.458 | 0.220 | 0.403 |
| mT0-large-ru | 0.842 | 0.680 | 0.752 | 0.813 | 0.800 | 0.805 | 0.443 | 0.232 | 0.414 |
| Enc2Enc | 0.811 | 0.588 | 0.682 | 0.800 | 0.768 | 0.782 | 0.393 | 0.164 | 0.343 |

The results presented in Table 4 demonstrate the effectiveness of various models finetuned on the persona extraction dataset. The majority of the models demonstrated high precision, rarely making mistakes. This may be due to the fact that most personas in the dialogue have a clear presence and, therefore, they are easily identified by almost all of the models. Additionally, models may extract a limited number of personas with which the model is highly confident, indicating that they belong to a specific dialogue participant. Consequently, the precision would be high.

However, it is worth noting that the models' low recall values indicate the difficulty in extracting all personas from the dialogues. This may be due to some personas being expressed implicitly, making it challenging for the models to recognize them. This is particularly evident in the comparison between Enc2Enc models and Bart, FRED-T5 or Starling in the table, where it can be observed that smaller models exhibit lower recall values.

Finetuned models without pretraining on similar tasks produce comparable results to pretrained models. This can be confirmed by comparing the F-measure and other metrics, which yield similar outcomes. Therefore, it can be concluded that pretraining the model is not essential to achieve the highest quality in the persona extraction task.

In relation to the hypothesis of using NCE Loss, it can be concluded that it affects the precision of both the MBart and FRED-T5 models. The precision value for FRED-T5 varies by approximately 2%, while for MBart, the difference is nearly 1%. However, it is important to note that the use of NCE Loss results in a decrease in recall. This may be because the models prioritize precision over generating all possible dialogue personas, possibly neglecting some of the target ones.

The finetuned RoBERTa classifier-based Enc2Enc model yields the lowest metrics. It struggles with generating personas, resulting in a low recall value. However, if speed or precision in extracting personas is a priority, these models may be useful in real-world tasks.

The FRED-T5 and Starling models, which are the largest among other models, demonstrated the most promising results. This suggests that as the size of a model increases, so does its ability to extract personas.

The utilization of translated and original samples is beneficial for enhancing the model's quality. This is evidenced by the observation that the metrics for mT0 and Starling, which were finetuned on such samples, exhibited higher values than those obtained in the absence of original samples. Furthermore, the metrics for mT0 exhibited a notable improvement. The low metrics value observed for ruT5 suggests that the model's multilinguality could potentially enhance its quality in instances where the data is constrained by translations.

It should also be noted that the quality of the persona classifier trained on the English-language persona dataset remains high when applied to Russian-language data. The knowledge acquired in English can be successfully transferred to the Russian language. The simplicity of the classification task required by the model may explain why it does not require syntactic or other language knowledge. The model only needs specific key phrases for accurate classification, which are identical in both Russian and English.

## 6 Discussion

### 6.1 Proposed Metrics Analysis

The comparison demonstrates that traditional metrics, such as rouge, bleu, bertscore, and so forth, fail to provide detailed insights into the shortcomings of the models. Proposed metrics, however, offer a means of identifying the challenges that the models face. Nevertheless, it remains unclear to what extent the metrics accurately reflect the quality of the models. To investigate this, we randomly selected a few samples and manually annotated them. Manual annotation was expected to reveal the prevalent issues and the actual model's quality.

Manual annotation was conducted in a manner analogous to automatic annotation. Instead of utilizing a similarity model and persona classifier, the extracted personas were manually matched with the target personas. This approach not only permitted the calculation of the metrics to be conducted correctly, but also enabled the typical errors associated with the proposed metrics to be evaluated. Table 5 shows the Pearson correlation between the count of manual and automatic calculation of matched personas described in Equations 1 and 2. Table 6 presents the manual and automatic metrics.

Table 5. The correlation between the automatic and manually annotated personas.

|  | Correlation |
|---|---|
| Total Extracted Personas | 1.000 |
| Total Target Personas + Classified Personas | 0.949 |
| Correctly Extracted Personas + Classified Personas | 0.874 |
| Matched Target Personas + Classified Personas | 0.737 |

Table 6. Metrics for manual and automatic annotations

| Annotation | P | R | F1 |
|---|---|---|---|
| Manual | 0.887 | 0.628 | 0.734 |
| Automatic | 0.858 | 0.722 | 0.784 |

There is a high correlation between manual and automatically counted personas. However, it is worth noting that the coefficient for Matched Relevant Instances is lower than all others. This, in turn, affects the recall metric. As can be seen in Table 6, automatic annotation slightly overestimates the recall value. Otherwise, the metrics of automatic annotation have almost the same values as those of manual annotation. This confirms the conclusions drawn about the quality of the models. Consequently, the proposed metrics could be employed in the evaluation of persona extraction.

### 6.2 Typical Metrics Errors

During the manual annotation process, we encountered several issues with the proposed metrics. One of the main challenges was the similarity model's tendency to mismatch sentences that express the presence or absence of specific objects. For instance, the personas "I don't have pets" and "I don't have a job" were incorrectly matched, despite both sentences indicating absence, but of different items. This highlights the limitation of the similarity model in capturing subtle differences in meaning.

Furthermore, the person extraction model may occasionally misidentify the subject of the persona, leading the similarity model to erroneously match the target and extracted personas. For instance, the model extracted the persona "I am reading Ender's Game" from the dialogue, whereas the target persona is "My son has finished the book Ender's Game." The similarity model matched both personas, which is incorrect because the actual persona refers to the son, not the subject.

Finally, there were difficulties in manual annotation of the extracted and target persona, particularly in situations where some personas are spoken of in the past tense. As an illustration, the following personas can be used: "I liked art" and "I don't like art". On the one hand, both personas can be identical in meaning, since the persona can be restated as "I used to like art, but now I don't". Conversely, it cannot be assumed that one does not currently dislike art, as this does not necessarily imply that one previously enjoyed it. Such circumstances introduce further complexity to the process of manual annotation.

To address some of the challenges, two approaches could be employed. Firstly, it is necessary to identify the threshold more accurately. Some personas were incorrectly matched due to a relatively high similarity value. Increasing the threshold could filter out such personas. Secondly, the E5 similarity model may not be the optimal choice. There are numerous multilingual similarity models that demonstrate high results on the MTEB. However, not all of them are suitable for matching personas. Consequently, it is necessary to conduct experiments to ascertain the most effective methodology for matching personas.

## 7 Conclusion

The study outlines a process for evaluating the effectiveness of various model training methods in extracting personas from dialogues. To assess the quality of these models, a metric based on F1-score has been developed, which considers both precision and recall of the model's predictions. Two algorithms have been developed for this purpose: one for matching targets with extracted information, and another for classifying each persona. The algorithms used in this study are based on the E5 sentence encoder and the persona classifier model.

A series of experiments were conducted utilizing a range of models, including the MBart, FRED-T5, Starling-7B based on the Mistral, mT0 and Enc2Enc models. Furthermore, the integration of Cross-Entropy and NCE Losses was employed to enhance the precision of persona extraction. The Starling-7B was identified as exhibiting the highest quality due to its substantial size. Significant quality improvement is achieved through the finetuning of models on translated and original examples. This approach facilitates cross-lingual knowledge transfer in a manner that is more effective than other approaches, including pretraining on similar tasks.

Despite the high precision of extracting personas, the research findings indicate that the models face challenges in extracting all personas from the dialogue, as evidenced by the low recall values obtained. This situation is observed even in large models such as Starling and Fred-T5. Therefore, it is not necessary to use LLMs for this task. In instances where computational resources are constrained, it may be feasible to utilize less costly models based on the Bart or T5 architectures, while maintaining an acceptable degree of precision.

**References**

1. Gibbeum Lee, Volker Hartmann, Jongho Park, Dimitris Papailiopoulos, and Kangwook Lee. 2023. Prompted LLMs as Chatbot Modules for Long Open-domain Conversation. In Findings of the Association for Computational Linguistics: ACL 2023, pp. 4536–4554, Association for Computational Linguistics, Toronto, Canada (2023).
2. Eric Chu, Prashanth Vijayaraghavan, and Deb Roy. Learning Personas from Dialogue with Attentive Memory Networks. In Proceedings of the 2018 Conference on Empirical Methods in Natural Language Processing, pp. 2638–2646, Association for Computational Linguistics, Brussels, Belgium (2018).
3. Wangchunshu Zhou, Qifei Li, and Chenle Li. Learning to Predict Persona Information for Dialogue Personalization without Explicit Persona Description. In Findings of the Association for Computational Linguistics: ACL 2023, pp. 2979–2991, Association for Computational Linguistics, Toronto, Canada (2023).

4. Ashish Vaswani, Noam Shazeer, Niki Parmar, Jakob Uszkoreit, Llion Jones, Aidan N. Gomez, Lukasz Kaiser, and Illia Polosukhin. 2017. Attention is all you need. In Advances in Neural Information Processing Systems 30: Annual Conference on Neural Information Processing Systems 2017, pp. 5998–6008, Long Beach, CA, USA, (2017).
5. Mike Lewis, Yinhan Liu, Naman Goyal, Marjan Ghazvininejad, Abdelrahman Mohamed, Omer Levy, Veselin Stoyanov, and Luke Zettlemoyer. BART: Denoising Sequence-to-Sequence Pre-training for Natural Language Generation, Translation, and Comprehension. In Proceedings of the 58th Annual Meeting of the Association for Computational Linguistics, pp. 7871–7880, Association for Computational Linguistics, Online (2020).
6. Colin Raffel, Noam Shazeer, Adam Roberts, Katherine Lee, Sharan Narang, Michael Matena, Yanqi Zhou, Wei Li, and Peter J. Liu. 2020. Exploring the limits of transfer learning with a unified text-to-text transformer. J. Mach. Learn. Res. 21, 1, Article 140 (January 2020), 67 pages.
7. Yinhan Liu, Jiatao Gu, Naman Goyal, Xian Li, Sergey Edunov, Marjan Ghazvininejad, Mike Lewis, and Luke Zettlemoyer. Multilingual Denoising Pre-training for Neural Machine Translation. Transactions of the Association for Computational Linguistics, vol. 8, pp. 726–742 (2020).
8. Linting Xue, Noah Constant, Adam Roberts, Mihir Kale, Rami Al-Rfou, Aditya Siddhant, Aditya Barua, and Colin Raffel. mT5: A Massively Multilingual Pre-trained Text-to-Text Transformer. In Proceedings of the 2021 Conference of the North American Chapter of the Association for Computational Linguistics: Human Language Technologies, pp. 483–498, Association for Computational Linguistics, Online (2021).
9. Niklas Muennighoff, Thomas Wang, Lintang Sutawika, Adam Roberts, Stella Biderman, Teven Le Scao, M Saiful Bari, Sheng Shen, Zheng Xin Yong, Hailey Schoelkopf, Xiangru Tang, Dragomir Radev, Alham Fikri Aji, Khalid Almubarak, Samuel Albanie, Zaid Alyafeai, Albert Webson, Edward Raff, and Colin Raffel. Crosslingual Generalization through Multitask Finetuning. In Proceedings of the 61st Annual Meeting of the Association for Computational Linguistics, pp. 15991–16111, Association for Computational Linguistics, Toronto, Canada (2023).
10. A Family of Pretrained Transformer Language Models for Russian. Dmitry Zmitrovich, Alexander Abramov, Andrey Kalmykov, Maria Tikhonova, Ekaterina Taktasheva, Danil Astafurov, Mark Baushenko, Artem Snegirev, Tatiana Shavrina, Sergey Markov, Vladislav Mikhailov, Alena Fenogenova. arXiv preprint arXiv: 2309.10931 (2023).
11. Patrick Lewis, Ethan Perez, Aleksandra Piktus, Fabio Petroni, Vladimir Karpukhin, Naman Goyal, Heinrich Küttler, Mike Lewis, Wen-tau Yih, Tim Rocktäschel, Sebastian Riedel, and Douwe Kiela. Retrieval-augmented generation for knowledge-intensive NLP tasks. In Proceedings of the 34th International Conference on Neural Information Processing Systems (NIPS'20). Article 793, pp. 9459–9474, Curran Associates Inc., Red Hook, NY, USA (2020).
12. Retrieval-Augmented Generation for Large Language Models: A Survey. Yunfan Gao, Yun Xiong, Xinyu Gao, Kangxiang Jia, Jinliu Pan, Yuxi Bi, Yi Dai, Jiawei Sun, Qianyu Guo, Meng Wang, Haofen Wang. arXiv preprint arXiv: 2312.10997 (2023).
13. Saizheng Zhang, Emily Dinan, Jack Urbanek, Arthur Szlam, Douwe Kiela, and Jason Weston. Personalizing Dialogue Agents: I have a dog, do you have pets too?. In Proceedings of the 56th Annual Meeting of the Association for Computational Linguistics. vol. 1, pp. 2204–2213, Association for Computational Linguistics, Melbourne, Australia (2018).
14. Jing Xu, Arthur Szlam, and Jason Weston. Beyond Goldfish Memory: Long-Term Open-Domain Conversation. In Proceedings of the 60th Annual Meeting of the Association for Computational Linguistics. vol. 1, pp. 5180–5197, Association for Computational Linguistics, Dublin, Ireland (2022).
15. Bogdan Gliwa, Iwona Mochol, Maciej Biesek, and Aleksander Wawer. SAMSum Corpus: A Human-annotated Dialogue Dataset for Abstractive Summarization. In Proceedings of the 2nd Workshop on New Frontiers in Summarization, pp. 70–79, Association for Computational Linguistics, Hong Kong, China (2019).
16. Yulong Chen, Yang Liu, Liang Chen, and Yue Zhang. DialogSum: A Real-Life Scenario Dialogue Summarization Dataset. In Findings of the Association for Computational Linguistics: ACL-IJCNLP 2021, pp. 5062–5074, Association for Computational Linguistics, Online (2021).
17. Daryna Dementieva, Daniil Moskovskiy, David Dale, and Alexander Panchenko. Exploring Methods for Cross-lingual Text Style Transfer: The Case of Text Detoxification. In Proceedings of the 13th International Joint Conference on Natural Language Processing and the 3rd Conference of the Asia-Pacific Chapter of the Association for Computational Linguistics. vol. 1, pp. 1083–1101, Association for Computational Linguistics, Nusa Dua, Bali (2023).
18. No Language Left Behind: Scaling Human-Centered Machine Translation. Marta R. Costa-jussà, James Cross, Onur Çelebi, Maha Elbayad, Kenneth Heafield, Kevin Heffernan, Elahe Kalbassi, Janice Lam, Daniel Licht, Jean Maillard, Anna Sun, Skyler Wang, Guillaume Wenzek, Al Youngblood, Bapi Akula, Loic Barrault, Gabriel Mejia Gonzalez, Prangthip Hansanti, John Hoffman, Semarley Jarrett, Kaushik Ram Sadagopan, Dirk Rowe, Shannon Spruit, Chau Tran, Pierre Andrews, Necip Fazil Ayan, Shruti Bhosale, Sergey Edunov, Angela Fan, Cynthia Gao, Vedanuj Goswami, Francisco Guzmán, Philipp Koehn, Alexandre Mourachko, Christophe Ropers, Safiyyah Saleem, Holger Schwenk, Jeff Wang. arXiv preprint arXiv: 2207.04672 (2022).
19. Vladislav Mikhailov, Tatiana Shamardina, Max Ryabinin, Alena Pestova, Ivan Smurov, and Ekaterina Artemova. RuCoLA: Russian Corpus of Linguistic Acceptability. In Proceedings of the 2022 Conference on Empirical Methods in Natural Language Processing, pp. 5207–5227, Association for Computational Linguistics, Abu Dhabi, United Arab Emirates (2022).
20. Irina Proskurina, Ekaterina Artemova, and Irina Piontkovskaya. Can BERT eat RuCoLA? Topological Data Analysis to Explain. In Proceedings of the 9th Workshop on Slavic Natural Language Processing 2023 (SlavicNLP 2023), pp. 123–137, Association for Computational Linguistics, Dubrovnik, Croatia (2023).
21. Multilingual E5 Text Embeddings: A Technical Report. Liang Wang, Nan Yang, Xiaolong Huang, Linjun Yang, Rangan Majumder, Furu Wei. arXiv preprint arXiv: 2402.05672 (2024).
22. Chin-Yew Lin. ROUGE: A Package for Automatic Evaluation of Summaries. In Text Summarization Branches Out, pp. 74–81, Association for Computational Linguistics, Barcelona, Spain (2004).
23. Kishore Papineni, Salim Roukos, Todd Ward, and Wei-Jing Zhu. Bleu: a Method for Automatic Evaluation of Machine Translation. In Proceedings of the 40th Annual Meeting of the Association for Computational Linguistics, pp. 311–318, Association for Computational Linguistics, Philadelphia, Pennsylvania, USA (2002).
24. Satanjeev Banerjee and Alon Lavie. METEOR: An Automatic Metric for MT Evaluation with Improved Correlation with Human Judgments. In Proceedings of the ACL Workshop on Intrinsic and Extrinsic Evaluation Measures for Machine Translation and/or Summarization, pp. 65–72, Association for Computational Linguistics, Ann Arbor, Michigan (2005).

25. Tianyi Zhang, Varsha Kishore, Felix Wu, Kilian Q. Weinberger, and Yoav Artzi. BERTScore: Evaluating Text Generation with BERT. In International Conference on Learning Representations (2020).
26. Zhu, B., Frick, E., Wu, T., Zhu, H., Ganesan, K., Chiang, W.L., Zhang, J., and Jiao, J. Starling-7B: Improving LLM Helpfulness & Harmlessness with RLAIF (2023).
27. Mistral 7B. Albert Q. Jiang, Alexandre Sablayrolles, Arthur Mensch, Chris Bamford, Devendra Singh Chaplot, Diego de las Casas, Florian Bressand, Gianna Lengyel, Guillaume Lample, Lucile Saulnier, Lélio Renard Lavaud, Marie-Anne Lachaux, Pierre Stock, Teven Le Scao, Thibaut Lavril, Thomas Wang, Timothée Lacroix, William El Sayed. arXiv preprint arXiv: 2310.06825 (2023).
28. Jörg Tiedemann. Parallel Data, Tools and Interfaces in OPUS. In Proceedings of the Eighth International Conference on Language Resources and Evaluation (LREC'12), pp. 2214–2218, European Language Resources Association (ELRA), Istanbul, Turkey (2012).
29. Representation Learning with Contrastive Predictive Coding. Aaron van den Oord, Yazhe Li, Oriol Vinyals. arXiv preprint arXiv: 1807.03748 (2019).
30. Sascha Rothe, Shashi Narayan, and Aliaksei Severyn. Leveraging Pre-trained Checkpoints for Sequence Generation Tasks. vol. 8, pp. 264–280, Transactions of the Association for Computational Linguistics (2020).
31. E. J Hu, Y. Shen, P. Wallis, Z. Allen-Zhu, Y. Li, S. Wang, L. Wang, W. Chen. LoRA: Low-Rank Adaptation of Large Language Models. In International Conference on Learning Representations (2022).

# Appendix A

| | Russian | English |
|---|---|---|
| Dialogue | bot_0: Привет, как ты сегодня?<br>bot_1: Отлично, спасибо! Только встаю. Поздно лег спать.<br>bot_0: Тогда доброе утро! Достаточно жарко для тебя? Мы не можем дождаться зимы.<br>bot_1: Да, мне нравится зима, хотя я и так не часто выхожу на улицу.<br>bot_0: То же самое и здесь, с моим инвалидным креслом немного тяжеловато.<br>bot_1: О, я так думаю! У меня нет инвалидной коляски, но я в школе. Компьютерная инженерия.<br>bot_0: Интересно! Приближается холодная погода, нужно быть осторожным с простудой. Не хочу пропускать занятия!<br>bot_1: Да, это правда. У меня настоящая страсть к компьютерному программированию.<br>bot_0: Я всегда запасаюсь витамином С, на всякий случай. Программирование звучит весело!<br>bot_1: Это весело. Я надеюсь использовать его, чтобы открыть свою собственную компанию с моим лучшим другом.<br>bot_0: Друзья - самые лучшие! Мой костюм купил мне машину в прошлом году. Совершенно меня удивило!<br>bot_1: Это потрясающе! Мой лучший друг на самом деле гей, а я нет.<br>bot_0: Любовь - это любовь! Удачи в вашем бизнесе!<br>bot_1: Я согласен! И большое спасибо. Надеюсь, все получится.<br>bot_0: Просто придерживайся этого, одной вещи, которой научила меня моя инвалидность, - это никогда не прекращать пытаться!<br>bot_1: Это очень вдохновляет, спасибо. | bot_0: Hi, how are you today?<br>bot_1: Fine, thanks! Just getting up, though. Went to bed late.<br>bot_0: Good morning, then! Hot enough for you? Ca not wait for winter.<br>bot_1: Yeah, I like winter, though I do not go outside much anyway.<br>bot_0: Same here, its a little hard with my wheelchair.<br>bot_1: Oh, I imagine so! I do not have a wheelchair, but I'm in school. Computer engineering.<br>bot_0: Interesting! With cooler weather coming you gotta watch out for colds. Don't wanna miss class!<br>bot_1: Yeah, that is true. I've a real passion for computer programming.<br>bot_0: I always stock up on vitamin c, just in case. Programming sounds like fun!<br>bot_1: Its fun. I hope to use it to open my own company with my best friend.<br>bot_0: Friends are the best! My vestie actually bought me a car last year. Totally surprised me!<br>bot_1: That's awesome! My best friend is actually gay, but I'm not.<br>bot_0: Love is love! Good luck with your business venture!<br>bot_1: I agree! And thanks very much. I hope it works out.<br>bot_0: Just keep to it one thing my disability has taught me is to never stop trying!<br>bot_1: That's extremely inspirational, thanks. |
| Persona | Мне нравится программировать компьютеры | I like computer programming |
| Target | 1 | 1 |

# Appendix B

| | Russian | English |
|---|---|---|
| Prompt | bot_0: Здравствуйте. Привет. Меня зовут Майк. - Как ты? - Хорошо.<br>bot_1: - Привет. - Здравствуйте. Я в порядке. - Как ты? - Хорошо.<br>bot_0: - Неплохо. Я только что вернулся из бассейна. Я люблю плавать.<br>bot_1: Мило! Я люблю собак, ну, вообще всех животных.<br>bot_0: Я тоже! У меня два кота и собака.<br>bot_1: Ух ты, прикольно! Я готова вернуться и закончить ветеринарную школу.<br>bot_0: Моя школа скоро начнётся. Нам многое предстоит прочитать, но я люблю читать.<br>bot_1: - Очень мило. Я тоже люблю читать. Какие книги тебе нравятся?<br>bot_0: Моя любимая - научная фантастика, но мне также нравятся книги по философии.<br>bot_1: - Очень мило. Я много читал о веганах до того, как стал одним из них.<br>bot_0: Эй, я тоже веган! Я очень высокий и моя кожа голубая, потому что я такой здоровый.<br>bot_1: Прекрасно! Твоя кожа голубая? Я не знаю.<br>bot_0: Да, немного. Я похож на персонажа из фильма "Аватар".<br>bot_1: Или смурфик. Мне понравился этот мультфильм.<br><br>Facts about bot_0: | bot_0: Hi. My name is mike. How are you?<br>bot_1: Hey. I'm good. How are you?<br>bot_0: Not bad. I just got back from the pool I love swimming.<br>bot_1: Nice! I have a love for dogs, well all animals really.<br>bot_0: Me too! I've two cats and a dog.<br>bot_1: Wow nice! I am ready to go back to finish vet school.<br>bot_0: My school is starting soon. We have a lot to read but I love reading.<br>bot_1: Nice. I also love to read. What types of books do you like?<br>bot_0: My favorite is science fiction but I also like philosophy books.<br>bot_1: Nice. I read a lot about vegan before becoming one.<br>bot_0: Hey I'm vegan too! I'm very tall and my skin is blue because I'm so healthy.<br>bot_1: Nice! Your skin is blue?!?!<br>bot_0: Haha yes a little bit. I look like a character from the avatar movie.<br>bot_1: Or a smurf. I loved that cartoon.<br><br>Facts about bot_0: |
| Target | Меня зовут Майк. Я люблю плавать. У меня два кота и собака. Я хожу в школу и люблю читать. Мне нравятся книги по научной фантастике и философии. Я также высокий и здоровый. Я веган. У меня голубоватая кожа. | My name is Mike. I love swimming. I have 2 cats and a dog. I go to school and love reading. I like science fiction and philosophy books. I'm also tall and healthy. I'm a Vegan. My skin is bluish. |